\documentclass[runningheads]{llncs}
\usepackage{graphicx}
\usepackage[labelformat=simple]{subcaption}
\usepackage{todonotes}
\graphicspath{ {images/} }
\usepackage{amssymb, amsmath}
\usepackage{tikz}
\usetikzlibrary{arrows,positioning,fit,shapes.geometric}
\tikzset{>=latex}

\begin{document}
\title{Demonstrating the Evolution of GANs through t-SNE}
\author{Victor Costa \and Nuno Louren\c{c}o \and Jo\~{a}o Correia \and Penousal Machado}
\institute{CISUC, Department of Informatics Engineering \\ University of Coimbra, Coimbra, Portugal\\\email{\{vfc, naml, jncor, machado\}@dei.uc.pt}}
\maketitle

\begin{abstract}
Generative Adversarial Networks (GANs) are powerful generative models that achieved strong results, mainly in the image domain. However, the training of GANs is not trivial, presenting some challenges tackled by different strategies. Evolutionary algorithms, such as COEGAN, were recently proposed as a solution to improve the GAN training, overcoming common problems that affect the model, such as vanishing gradient and mode collapse. In this work, we propose an evaluation method based on t-distributed Stochastic Neighbour Embedding (t-SNE) to assess the progress of GANs and visualize the distribution learned by generators in training. We propose the use of the feature space extracted from trained discriminators to evaluate samples produced by generators and from the input dataset. A metric based on the resulting t-SNE maps and the Jaccard index is proposed to represent the model quality. Experiments were conducted to assess the progress of GANs when trained using COEGAN. The results show both by visual inspection and metrics that the Evolutionary Algorithm gradually improves discriminators and generators through generations, avoiding problems such as mode collapse.

\end{abstract}

\section{Introduction}
Generative Adversarial Networks (GANs)~\cite{NIPS2014_5423} gained relevance in the past years for producing impressive results in the context of images.
The GAN model uses adversarial training to achieve strong discriminative and generative components.
The typical model is comprised of two neural networks: a generator and a discriminator.
These networks compete in a unified training process where the generator uses its neural network to produce samples and the discriminator tries to classify these samples as fake or real (i.e., drawn from the input dataset).
Although the discriminator trained by a GAN also represents an important outcome of the training process, GANs are mostly used as a generative model to produce innovative samples based on an input distribution.

Despite the progress regarding generative models, the training of GANs is challenging and is affected by some well-known issues, such as the vanishing gradient and mode collapse~\cite{brock2018large,fedus2017many}.
Therefore, a trial-and-error approach is usually applied to obtain the expected results, making the training of GANs an uncertain process.
The vanishing gradient and the mode collapse problems are related to the balance between the discriminator and the generator.
Vanishing gradient occurs when the discriminator or generator becomes much more powerful than the other, leading to stagnation of training.
Mode collapse occurs when the generator captures only a small fraction of the input distribution.

Several improvements were proposed for the initial GAN model.
These proposals aim not only to improve the GAN training but also to produce more realistic results, focusing on two aspects of GANs: loss functions and architectural mechanisms of the neural networks.
Therefore, more efficient loss functions and architectural improvements were proposed, such as WGAN~\cite{arjovsky2017wasserstein}, BEGAN~\cite{berthelot2017began}, LSGAN~\cite{mao2017least}, SN-GAN~\cite{miyato2018spectral}, StyleGAN~\cite{karras2018style}, and DCGAN~\cite{radford2015unsupervised}.
However, stability issues in GAN training are still present.

Neuroevolution is an approach used to design and optimize neural networks through the application of evolutionary algorithms~\cite{miikkulainen2017evolving,neat,yao1999evolving}.
These algorithms are based on the evolutionary mechanism found in nature, evolving a population of individuals through selective pressure, leading to the discovery of efficient solutions for a certain problem~\cite{sims1994evolving}.
Recently, a combination of evolutionary algorithms and GANs were proposed to improve the original model.
Progress was made in both the resulting quality of the outcome and the stability of the GAN training.
Methods such as E-GAN~\cite{wang2018evolutionary}, Pareto GAN~\cite{garciarena2018evolved}, Lipizzaner~\cite{al2018towards}, Mustangs~\cite{toutouh2019spatial}, and COEGAN~\cite{costa2019evaluating,costa2019coevolution} use different approaches to apply Evolutionary Algorithms on the training of GANs.

Coevolutionary GAN (COEGAN)~\cite{costa2019evaluating,costa2019coevolution} combines neuroevolution and coevolution on the orchestration of the GAN training.
Namely, competitive coevolution is used to design the algorithm in order to produce evolutionary pressure and overcome the stability issues affecting the training of GANs.
The authors showed through experimental analysis that the method was able to discover efficient models for GANs in different datasets~\cite{costa2019evaluating}.

We propose in this paper a new method to evaluate the progress of GANs during the training process.
Therefore, we design an evaluation method that uses t-distributed Stochastic Neighbour Embedding (t-SNE)~\cite{maaten2008visualizing} to visualize and quantify the performance of discriminators and generators during the evolutionary process.
For this, we use the feature space produced by discriminators to analyze images produced by generators and drawn from the input dataset.
The t-SNE algorithm was fed with this feature space in order to distribute those images in a two-dimensional grid.
A metric based on the Jaccard index in the resulting t-SNE maps was proposed to quantify the performance achieved by GAN models.

This evaluation method was applied to analyze the evolution of discriminators and generators in COEGAN.
The experiments evidenced that the distribution of samples produced by our evaluation method is able to create a consistent visualization of the evolutionary process in COEGAN.
The results provide additional evidence of the evolution of generators and discriminators achieved by COEGAN.

The remainder of this paper is organized as follows:
Section \ref{sec:background} introduces t-SNE, GANs, and neuroevolution, presenting state-of-the-art works using these concepts;
Section \ref{sec:model} summarizes the COEGAN algorithm;
Section \ref{sec:evaluation} presents the method proposed in this work to evaluate the progress of discriminators and generators;
Section \ref{sec:experiments} displays the experimental results of COEGAN using our evaluation method;
finally, Section \ref{sec:conclusions} presents our conclusions and future work.

\section{Background and Related Works}
\label{sec:background}
In this section, we present the concepts used in this work to develop the evaluation method and apply it in Evolutionary Algorithms.
Therefore, we will introduce concepts of neuroevolution and GANs.
We will also describe works proposed to use Evolutionary Algorithms in the training of GANs.
Finally, we describe the t-SNE algorithm and show works using it to represent and evaluate data distributions.

\subsection{Neuroevolution}
Evolutionary Algorithms find inspiration in nature to design mechanisms based on biological evolution~\cite{sims1994evolving}.
Several strategies to apply these evolution mechanisms were proposed in the literature, offering new perspectives to solve a variety of problems.
In general, Evolutionary Algorithms use a population of potential solutions to solve a defined problem, using variation operators and selective pressure to adapt individuals toward the target.
Thus, each solution is an individual represented through an abstraction called genotype.
The genotype transformation derives the concrete solution, called phenotype.

Neuroevolution is the application of Evolutionary Algorithms to the evolution of neural networks.
In this case, the genotype represents an abstraction for the implementation of a neural network.
This representation can be direct, i.e., all nodes and connections of the neural architecture are encoded~\cite{miikkulainen2017evolving,neat}, or indirect, i.e., rules are specified to derive the concrete implementation of neural networks, such as in structured grammatical evolution~\cite{assunccao2019denser,lourencco2015sge}.

Weights, topology, and hyperparameters can be evolved through neuroevolution~\cite{yao1999evolving}.
Thus, the manual process used by researchers to discover efficient models can be transformed into an automatic process.
It is important to note that the training of neural networks is a time-consuming task, impacting the performance of neuroevolution algorithms.

NeuroEvolution of Augmenting Topologies (NEAT)~\cite{neat} is a well-known model that uses neuroevolution in the evolution of both weights and topologies of neural networks.
The genotype is a direct representation of the neural network, where NEAT defines two lists for the genome of individuals: a list of neurons and a list of connections between these neurons.
A further expansion of NEAT was proposed to enable larger search spaces in DeepNEAT and CoDeepNEAT~\cite{miikkulainen2017evolving}.
In these models, the genes composing a genome are abstractions of entire layers, enabling the representation of deep neural networks.
Deep neural networks became popular and achieved strong performance in several tasks.
Thus, the need for automation became relevant to improve progress on deeper models~\cite{miikkulainen2017evolving}.

\subsection{Generative Adversarial Networks}
In Generative Adversarial Networks (GANs)~\cite{NIPS2014_5423}, two neural networks are used in an adversarial way in a unified training process.
These networks are represented by one generator and one discriminator.
The discriminator is trained with some input dataset and has to classify samples as originated from this dataset (i.e., real samples) or samples produced by the generator (i.e., fake samples).
Thus, the discriminator outputs a probability of each sample belonging to the input dataset.
On the other hand, the generator receives a probability distribution as input and transforms this input into fake samples similar to the input dataset.
These fake samples are also used as input to the discriminator to evaluate their quality.
This adversarial model leads to the creation of strong generative and discriminative models, taking advantage of these adversarial characteristics to progressively improve their performance.
Generators trained by GANs produce high-quality results without losing the characteristics of the input data.

\begin{figure}
	\centering
	\begin{tikzpicture}[
	node distance=7mm,
	title/.style={font=\fontsize{6}{6}\color{black!50}\ttfamily},
	block/.style= {draw, rectangle, align=center,minimum width=2.5cm,minimum height=1cm},
	loss/.style= {draw, trapezium, align=center,minimum width=1cm,minimum height=0.3cm},
	image/.style= {align=center,minimum width=1cm,minimum height=1cm,inner sep=2pt},
	typetag/.style={rectangle, draw=black!50, font=\scriptsize\ttfamily, anchor=west}
	]
	\node [image] (input) {\includegraphics[width=.08\textwidth]{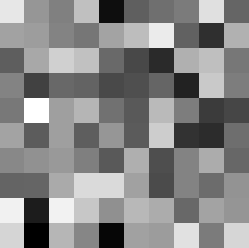}};
	\node [text width=2cm, below =0cm of input] {random input};
	\node [block, right =1cm of input, line width=0.8mm]  (generator) {Generator};
	\node [loss, below =-0.01cm of generator, shape border rotate=180]  (lossg) {loss};
	\node [image, right =1cm of generator]  (sample) {\includegraphics[width=.08\textwidth]{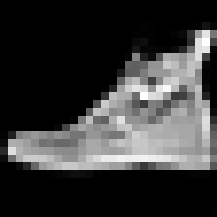}};
	\node [text width=2.2cm, below =0cm of sample] (sample-text) {created sample};
	\node [block, below =1.0cm of lossg, line width=0.8mm]  (discriminator) {Discriminator};
	\node [loss, above =-0.01cm of discriminator]  (lossd) {loss};
	\node [image, left =1cm of discriminator] (dataset) {\includegraphics[width=.15\textwidth]{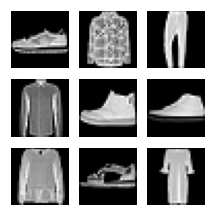}};
	\node [text width=1.9cm, below =0cm of dataset] {input dataset};

	\path[draw,->]
	(input) edge (generator)
	(generator) edge (sample)
	(sample-text) |- (discriminator)
	(dataset) edge (discriminator)
	(lossd) edge (lossg)
	;
	\end{tikzpicture}
	\caption{Interaction between the components of a GAN during the training process.} \label{fig:gan}
\end{figure}
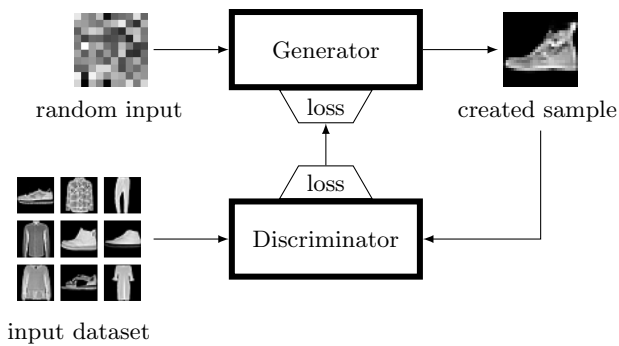

Figure~\ref{fig:gan} describes the interaction between discriminators and generators in a GAN when using a dataset such as Fashion MNIST~\cite{xiao2017fashion}.

The GAN training algorithm works iteratively.
At each iteration, backpropagation is applied for training the discriminator and the generator with their respective loss functions in order to update their network parameters.
The loss function of the discriminator ($\mathcal{L}_D^{GAN}$) is defined as follows:
\begin{equation}
\mathcal{L}_D^{GAN} = -\mathbb{E}_{x \sim p_d}[\log D(x)] - \mathbb{E}_{z \sim p_z}[\log(1 - D(G(z)))].
\label{eq:discriminator}
\end{equation}

For the generator, the non-saturating version of the loss function ($\mathcal{L}_G^{GAN}$) is defined by:
\begin{equation}
\mathcal{L}_G^{GAN} = - \mathbb{E}_{z \sim p_z}[\log(D(G(z)))].
\label{eq:generator}
\end{equation}

In Eq.~\eqref{eq:discriminator}, $p_d$ represents the input dataset and $x$ represents its samples.
In Eq.~\eqref{eq:discriminator} and Eq.~\eqref{eq:generator}, $p_z$ is the distribution used for the generator, $z$ is the latent space drawn from $p_z$, $G$ is the generator, and $D$ represents the discriminator.

GANs lead to relevant progress for generative models.
However, the training of GANs is challenging, and stability issues frequently affect the process.
Two common issues that impact training are the vanishing gradient and the mode collapse problem.
The vanishing gradient occurs when the capacities of the generator and the discriminator are not in equilibrium, making one of them too powerful when compared to the other.
Thus, the gradient does not properly improve the performance and the training progress stagnates.
The mode collapse issue occurs when the generator fails to capture the input distribution used on training.
In this case, samples created by the generator will partially represent this distribution.
For example, the mode collapse occurs in a digits dataset when only some digits are represented in the set of created samples.

The Fr\'{e}chet Inception Distance (FID)~\cite{heusel2017gans} is frequently used to evaluate the performance of GANs.
FID uses the outcome of the last hidden layer in Inception Net~\cite{szegedy2016rethinking} (trained on ImageNet~\cite{russakovsky2015imagenet}) to transform images from the input dataset and created by generators into a feature space.
This feature space is interpreted as a continuous multivariate Gaussian, and the mean and covariance of the two resulting Gaussians are used to calculate the Fr\'{e}chet distance as:
\begin{equation}
FID(x,g) = ||\mu_x - \mu_g||_2^2 + Tr(\varSigma_x + \varSigma_g - 2(\varSigma_x\varSigma_g)^{1/2}),
\label{eq:fid}
\end{equation}
with $\mu_x$, $\varSigma_x$, $\mu_g$, and $\varSigma_g$ representing the mean and covariance estimated for the input dataset $x$ and fake samples $g$, respectively.
FID is able to quantify the quality and diversity of the generative model, outperforming other metrics such as the Inception Score~\cite{salimans2016improved}.

Several improvements over the original GAN model were proposed to minimize these issues and leverage the quality of the results.
In this context, new loss functions were proposed to replace the original losses (Eq.~\eqref{eq:discriminator} and Eq.~\eqref{eq:generator}), such as in WGAN~\cite{arjovsky2017wasserstein}, LSGAN~\cite{mao2017least}, and SN-GAN~\cite{miyato2018spectral}.
Besides loss functions, architectural improvements were proposed for the GAN model.
In DCGAN~\cite{radford2015unsupervised}, a reference architecture for the discriminator and the generator was proposed.
In~\cite{karras2018progressive}, the authors propose a predefined strategy to progressively grow a GAN during the training process.
SAGAN~\cite{zhang2018self} uses self-attention modules to model the relationship between spatial regions of the input sample.

These alternative loss functions and architectural improvements minimize some problems and produce better results, but issues still affect the training of GANs~\cite{arjovsky2017wasserstein,gulrajani2017improved,salimans2016improved}.
Besides, efficient models designed for a specific task are not guaranteed to work properly in other tasks.
Thus, the discovery of efficient models and hyperparameters is not trivial, requiring recurrent empirical validation depending on the underlying problem.

\subsection{Evolutionary Algorithms and GANs}
Recently, Evolutionary Algorithms were proposed to train and evolve GANs.
These solutions make use of different mechanisms of evolutionary computation not only to minimize stability problems in GANs but also to produce better outcomes concerning the quality of created samples.

E-GAN~\cite{wang2018evolutionary} proposes a variation operator that switches the loss functions of generators.
The architectures of generators and the discriminator are fixed and based on DCGAN~\cite{radford2015unsupervised}.
In~\cite{garciarena2018evolved}, neuroevolution was used in combination with Pareto set approximations to evolve GANs.
In this case, the architecture is not fixed and evolves through generations.
Lipizzaner~\cite{al2018towards} uses spatial coevolution to train GANs.
The architectures of generators and discriminators are fixed.
A mixture of weights is used to compose generators through an evolution strategy based on their spatial neighborhood.
In Mustangs~\cite{toutouh2019spatial}, the Lipizzaner model was extended to combine the E-GAN dynamic loss function with the spatial coevolution mechanism from Lipizzaner.
COEGAN~\cite{costa2019evaluating,costa2019coevolution} designs an algorithm that combines competitive coevolution and neuroevolution with the GAN training process.
We provide an overview of the method in Section~\ref{sec:model}.
A comparison between these algorithms can be seen in~\cite{costa2020neuroevolution}.

\subsection{t-SNE}
t-distributed Stochastic Neighbour Embedding (t-SNE)~\cite{maaten2008visualizing} is a technique used to produce a map (two or three dimensions) that represents the data distribution.
Therefore, t-SNE is useful to provide the visualization of complex distributions by revealing the structure of the data.
t-SNE was applied in a variety of problems from different fields~\cite{belkina2019automated,kobak2019art}.
It was also used to visualize the distribution of images produced by GANs in~\cite{zhang2018stackgan++}.

The t-SNE algorithm works iteratively.
A set of pairwise affinities is calculated for the input data and the solution is randomly initialized using a probabilistic distribution.
At each iteration, the gradient is calculated based on the Kullback-Leibler divergence between the high-dimensional input space and the corresponding lower-dimensional representation.
The gradient is used to update the solution.
After all iterations, t-SNE outputs the final solution, representing points in a two or three-dimensional grid for the input data.

The number of iterations and perplexity are two important parameters for the t-SNE algorithm.
Perplexity defines how the neighborhood of each data point is handled.
The range $[5, 50]$ is recommended for perplexity.
The number of iterations limits the number of steps used to update the final solution.

Principal Components Analysis (PCA) can be used as a preprocessing step to reduce the dimensionality of the data, suppress noise, and achieve a faster computation~\cite{kobak2019art,maaten2008visualizing}.
For example, PCA was used to reduce the dimensionality of the data to $30$ and $50$ in \cite{maaten2008visualizing} and \cite{kobak2019art}, respectively.

\section{COEGAN}~\label{sec:model}
Coevolutionary Generative Adversarial Networks (COEGAN) algorithm combines neuroevolution and competitive coevolution on the training and evolution of GANs~\cite{costa2019evaluating,costa2019coevolution}.
COEGAN was initially inspired by NEAT~\cite{neat} and DeepNEAT \cite{miikkulainen2017evolving} to develop a representation for neural networks.
Nevertheless, COEGAN adapts the evolutionary model to the context of GANs.
In this section, we describe the fundamental aspects of the algorithm.
More details of the algorithm can be found in~\cite{costa2019evaluating,costa2019coevolution}.

In COEGAN, a population of discriminators and another of generators are used in a competitive coevolution setup.
At each generation, generators and discriminators are paired for the application of the original GAN training algorithm, using backpropagation to learn internal weights and bias of neural networks.
COEGAN can be used with different pairing strategies.
One simple strategy is the \textit{all vs. all} pairing, which defines all possible pairs between generators and discriminators in the current population.
Another possibility is to take only the best individuals for pairing.
This strategy defines the \textit{all vs. $k$-best} pairing, using the top $k$ individuals from each population for the GAN training.
For performance reasons, each pair in COEGAN is trained with a limited set of data at each generation.
However, COEGAN reuses the weights and bias of individuals in the breeding process, resembling a transfer learning mechanism to obtain a full representation of the input distribution.
The fitness of each individual is derived from the results of these matches between generators and discriminators.

\subsection{Representation}
In COEGAN, individuals use a genotype composed of a sequential array of genes.
These genes are directly transformed into sequential layers in a neural network and can represent a linear (fully connected), convolution, or deconvolution (transpose convolution) layer.
Each type of gene has internal parameters, such as the activation function and the number of output features, that are subject to the variation operators.
Some internal parameters are dynamically defined at the phenotype transformation phase, making use of the setup of the genome to form a valid neural network for the application of backpropagation.
Therefore, the number of input channels, stride, and kernel size are parameters adjusted based on the configuration of the previous layer for convolution layers.

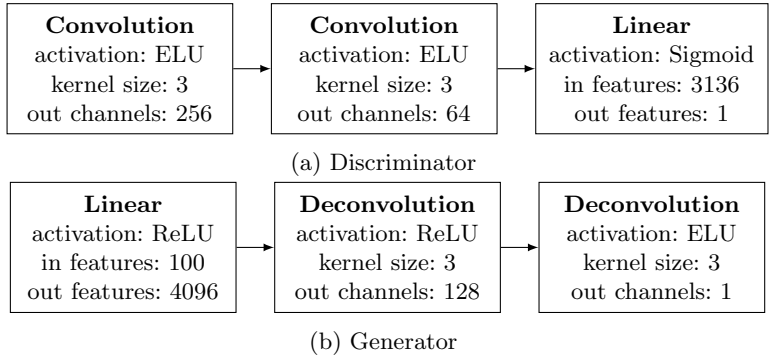
\begin{figure}
	\centering
	\subfloat[Discriminator]{\label{fig:discriminator_genotype}
	\begin{tikzpicture}[
	node distance=8mm,
	block/.style= {draw, rectangle, align=center,minimum width=3cm,minimum height=1cm, inner sep=5pt},
	]
	\node [block] (l1) {\textbf{Convolution}\\activation: ELU\\kernel size: 3\\out channels: 256};
	\node [block, right =0.5cm of l1] (l2) {\textbf{Convolution}\\activation: ELU\\kernel size: 3\\out channels: 64};
	\node [block, right =0.5cm of l2] (l3) {\textbf{Linear}\\activation: Sigmoid\\in features: 3136\\out features: 1};
	\path[draw,->]
	(l1) edge (l2)
	(l2) edge (l3)
	;
	\end{tikzpicture}
	}

	\subfloat[Generator]{\label{fig:generator_genotype}
	\begin{tikzpicture}[
	node distance=8mm,
	block/.style= {draw, rectangle, align=center,minimum width=3cm,minimum height=1cm, inner sep=5pt},
	]
	\node [block] (l1) {\textbf{Linear}\\activation: ReLU\\in features: 100\\out features: 4096};
	\node [block, right =0.5cm of l1] (l2) {\textbf{Deconvolution}\\activation: ReLU\\kernel size: 3\\out channels: 128};
	\node [block, right =0.5cm of l2] (l3) {\textbf{Deconvolution}\\activation: ELU\\kernel size: 3\\out channels: 1};
	\path[draw,->]
	(l1) edge (l2)
	(l2) edge (l3)
	;
	\end{tikzpicture}
	}
	\caption{Example of genotypes of a discriminator and a generator. The discriminator contains two convolution layers and one linear layer. The generator has one linear and two deconvolution layers. The parameters are listed for each gene (e.g., activation type, kernel size, and the number of channels).}
	\label{fig:genotype}
\end{figure}

Figure~\ref{fig:genotype} illustrates the genotypes of a discriminator and a generator.
In Figure~\ref{fig:discriminator_genotype}, the discriminator contains a convolutional section with two layers, followed by a linear section of a single output layer.
This output layer returns the probability of samples to be real or fake.
In Figure~\ref{fig:generator_genotype}, the generator starts with a linear layer followed by two deconvolutional layers.
The last layer returns synthetic samples with the same characteristics (i.e., shape and number of color channels) as samples from the input dataset.

\subsection{Fitness}
COEGAN uses specific fitness functions for discriminators and generators.
The fitness of discriminators is based on the loss function of the original GAN model (Eq.~\eqref{eq:discriminator}).
The fitness of generators is based on the FID score (Eq.~\eqref{eq:fid}).
These fitness functions were chosen to produce selection pressure on the evolution of better individuals.

\subsection{Selection}
The selection mechanism in COEGAN is based on the speciation originally proposed on NEAT~\cite{neat}.
In COEGAN, a distance function based on the genome similarity is used to subdivide each population into species.
Thus, individuals with similar neural networks have a tendency to belong to the same group.
On the other hand, individuals modified by mutation can form new species, making use of speciation to protect their innovation.

In the selection phase, COEGAN uses the speciation strategy to select individuals for asexual reproduction.
Individuals from each population are selected in proportion to the average fitness of their species.
Tournament is also applied inside each species to finally determine the survivors.

\subsection{Variation Operators}
COEGAN uses only mutations as variation operators.
Three types of mutations were defined with the purpose of adding, removing, and modifying layers.
The addition operator randomly initializes a new layer and inserts it into the genotype.
The removal operator randomly removes a layer from the genotype.
The change operator modifies the internal attributes of a randomly selected gene (e.g., the activation function, number of channels, and number of features).

An adaptive step is executed after reproduction to ensure the transference of the learned internal parameters between compatible individuals through generations.
Thus, when possible, COEGAN copies the weights and bias of the parent into the offspring.
However, parameters are reinitialized when architectures are incompatible.

\section{Evaluation Method}~\label{sec:evaluation}
We proposed in this work a new method to visualize and evaluate the progress of discriminators and generators in GANs.
We applied this method in COEGAN to provide further evidence of the evolutionary contribution of the model to the creation of strong generators and discriminators.
Nevertheless, this method can also be applied in regular GANs (e.g., the original GAN model or WGAN).

\begin{figure}
\centering
\begin{tikzpicture}[
	node distance=8mm,
	title/.style={font=\fontsize{8}{8}\color{black!80}\ttfamily},
	block/.style= {draw, rectangle, align=center,minimum width=2cm,minimum height=1cm},
	image/.style= {align=center,minimum width=2cm,minimum height=1cm,inner sep=0pt},
	typetag/.style={rectangle, draw=black!50, font=\scriptsize\ttfamily, anchor=west}
]
\node [title] (coegan) {COEGAN};
\node [block, below=of coegan.west, typetag]  (discriminator) {Discriminator};
\node [block, below =1cm of discriminator]  (generator) {Generator};
\node [draw=black!80, line width=0.8mm, inner sep=10pt, fit={(coegan) (discriminator) (generator)}] {};
\node [image, left =1cm of discriminator] (dataset) {\includegraphics[width=.12\textwidth]{input_samples.png}\\input dataset};
\node [block, right =1cm of discriminator] (pca) {PCA};
\node [block, right =1cm of pca, line width=0.8mm] (tsne) {t-SNE};
\node [image, below =0.5cm of tsne] (tsneimage) {\includegraphics[width=.2\textwidth]{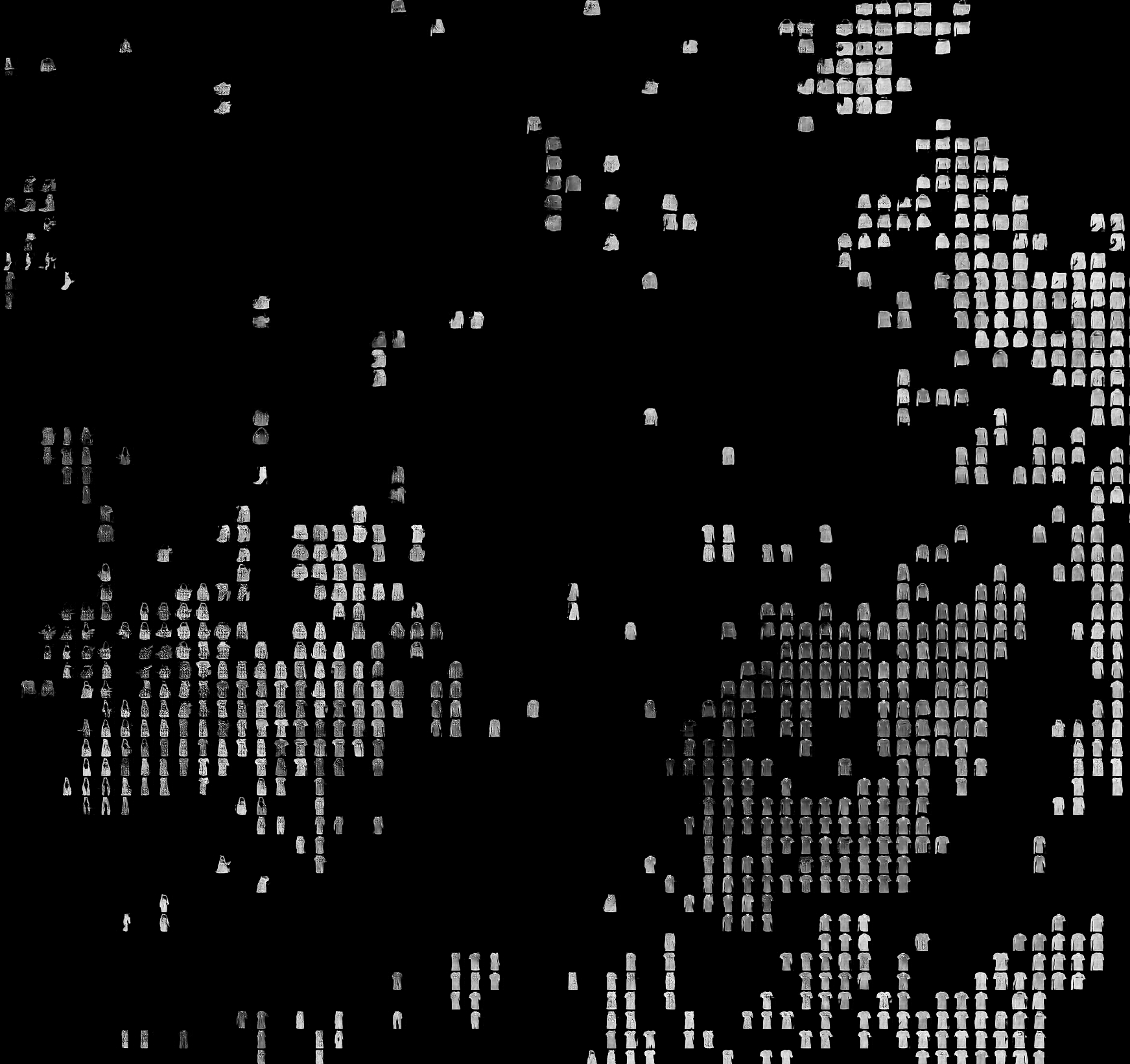}\\2d grid};
\node [block, left =0.6cm of tsneimage, line width=0.8mm] (metrics) {Metric:\\Jaccard Index};

\path[draw,->]
(discriminator) edge (pca)
(generator) edge (discriminator)
(dataset) edge (discriminator)
(pca) edge (tsne)
(tsne) edge (tsneimage)
(tsneimage) |- (metrics)
;
\end{tikzpicture}
\caption{Overview of the evaluation method proposed in this work to analyze the progress of generators and discriminators in GANs.} \label{fig:eval_method}
\end{figure}

Figure~\ref{fig:eval_method} presents an overview of the evaluation method.
We initially train COEGAN with the input dataset.
After training, we use snapshots of the best discriminators and generators from different generations to visualize the performance through t-SNE.
For the t-SNE calculation, we provide to discriminators samples from the input dataset and samples created by generators, using the output of the last hidden layer to construct a high-dimensional features matrix.
To improve the performance, we apply Principal Components Analysis (PCA) to reduce the number of features and use the resulting matrix to fed t-SNE.

This matrix contains data from the input dataset and also from all evaluated generators.
Thus, the resulting data of all inputs are jointly used for creating a lower-dimensional representation of the data through the t-SNE algorithm.
Then, the output of t-SNE is transformed into a two-dimensional grid that spatially distributes the input images.
This grid represents a map revealing the distribution of samples according to their inner characteristics.
Thus, we can visualize problems such as mode collapse by inspecting the grid and ensuring that the distribution of samples is not concentrated in a single region.
We can also visually compare the distribution of samples from the input dataset with samples from the generator to assess the completeness of the generative model.
Furthermore, by using discriminators to transform images into a feature space, we also assess their capacity to classify samples.

Besides these visualizations, we also propose a metric to quantify the performance of the model.
For this, given a map $M^G$ of samples produced by a generator (in a specific generation) and a map $M^d$ produced by the input dataset, we calculate the Euclidean distances ${D}_{(i,j)}$ between all samples in $M^G$ and $M^d$:
\begin{equation}
\mathcal{D}_{i,j} = \lVert M^G_i - M^d_j \lVert.
\label{eq:distances}
\end{equation}

Samples in $M^G$ and $M^d$ are not perfectly equal and distances in $\mathcal{D}_{i,j}$ are not zeroed.
Thus, we use these distances to define a threshold for the similarity between samples.
A global threshold $\tau$ is defined by the median of the minimum distances in $\mathcal{D}_{i}$ for maps $M^G$ related to the last generation.
This threshold defines the set of samples in $M_G$ with corresponding samples in $M_d$ as:
\begin{equation}
\mathcal{I}^G = \{ M^G_i |  \exists j, D_{i,j} < \tau \}.
\label{eq:intersection}
\end{equation}

The set $\mathcal{I}^G$ contains the samples in $M_G$ that were successfully approximated by a sample in $M_d$, evidencing that this part of the input distribution was captured by the model.
Thus, we consider this set as the intersection between these two grids and calculate the Jaccard index as:
\begin{equation}
J^G=\frac{|\mathcal{I}^G|}{|M^G \cup M^d|}
\label{eq:metric}
\end{equation}

In this work, we use Eq.\eqref{eq:metric} to quantify the quality of models.
A high $J^G$ indicates that the generator was able to capture the input distribution successfully.
On the other hand, a perfect score in this metric indicates that the generative model is not able to produce innovative samples.

\section{Experiments}~\label{sec:experiments}
Experiments were conducted to assess the evolution of generators and discriminators in COEGAN using the evaluation method proposed in this work.
The Fashion MNIST dataset was used in COEGAN training to gauge the characteristics of the proposed evaluation method.

\subsection{Experimental Setup}
Table \ref{table:setup} describes the parameters used in COEGAN, chosen based on previous experiments~\cite{costa2019evaluating,costa2019coevolution}.
We train COEGAN for $100$ generations in a population of generators and discriminators of $10$ individuals using the \textit{all vs. all} pairing strategy.
The probabilities for mutations to add, remove, or change genes are 30\%, 10\%, and 10\%, respectively.
The genome was limited to four genes, representing a network of four layers in the maximum allowed setup.
Only convolution and transpose convolution were used as options when creating new layers.
This setup was sufficient to discover efficient solutions in the experiments with Fashion MNIST.
We use three species in each population of generators and discriminators.
The FID score was applied with $5000$ samples to evaluate generators.

\begin{table}[!t]
	\renewcommand{\arraystretch}{1.3}
	\caption{Experimental Parameters}
	\label{table:setup}
	\centering
	\begin{tabular}{c c}
		\hline\hline
		\textbf{Evolutionary Parameters} & \textbf{Value} \\
		\hline
		Number of generations & 100 \\
		Population size (generators and discriminators) & $10$, $10$ \\
		Probabilities (add, remove, change) & 30\%, 10\%, 10\% \\
		Output channels range & [32, 512] \\
		Tournament $k_t$ & 2 \\
		FID samples & 5000 \\
		Genome Limit & 4 \\
		Species & 3 \\
		\textbf{GAN Parameters} & \textbf{Value} \\
		\hline
		Batch size & 64 \\
		Batches per generation & 10 \\
		Optimizer & Adam \\
		Learning rate & 0.003 \\
		\textbf{Evaluation Parameters} & \textbf{Value} \\
		\hline
		PCA dimensions & 50 \\
		t-SNE Perplexity & 30 \\
		t-SNE Iterations & 1000 \\
		Samples per model & 1000 \\
		\hline\hline
	\end{tabular}
\end{table}

For each training pair where the GAN training is applied, we use $10$ batches of $64$ images and the Adam optimizer~\cite{kingma2015adam} with $0.003$ as the learning rate.

In Table~\ref{table:setup}, we also list the parameters used in the evaluation method.
PCA was applied to reduce the dimensionality of the data to $50$.
For the t-SNE algorithm, $30$ and $1000$ were used as perplexity and number of iterations, respectively.
We use $1000$ samples for each model and from the input dataset to obtain the two-dimensional map through t-SNE.

\subsection{Results}
\begin{figure}[ht]
	\centering
	\subfloat[Generation 5]{\includegraphics[width=0.37\linewidth]{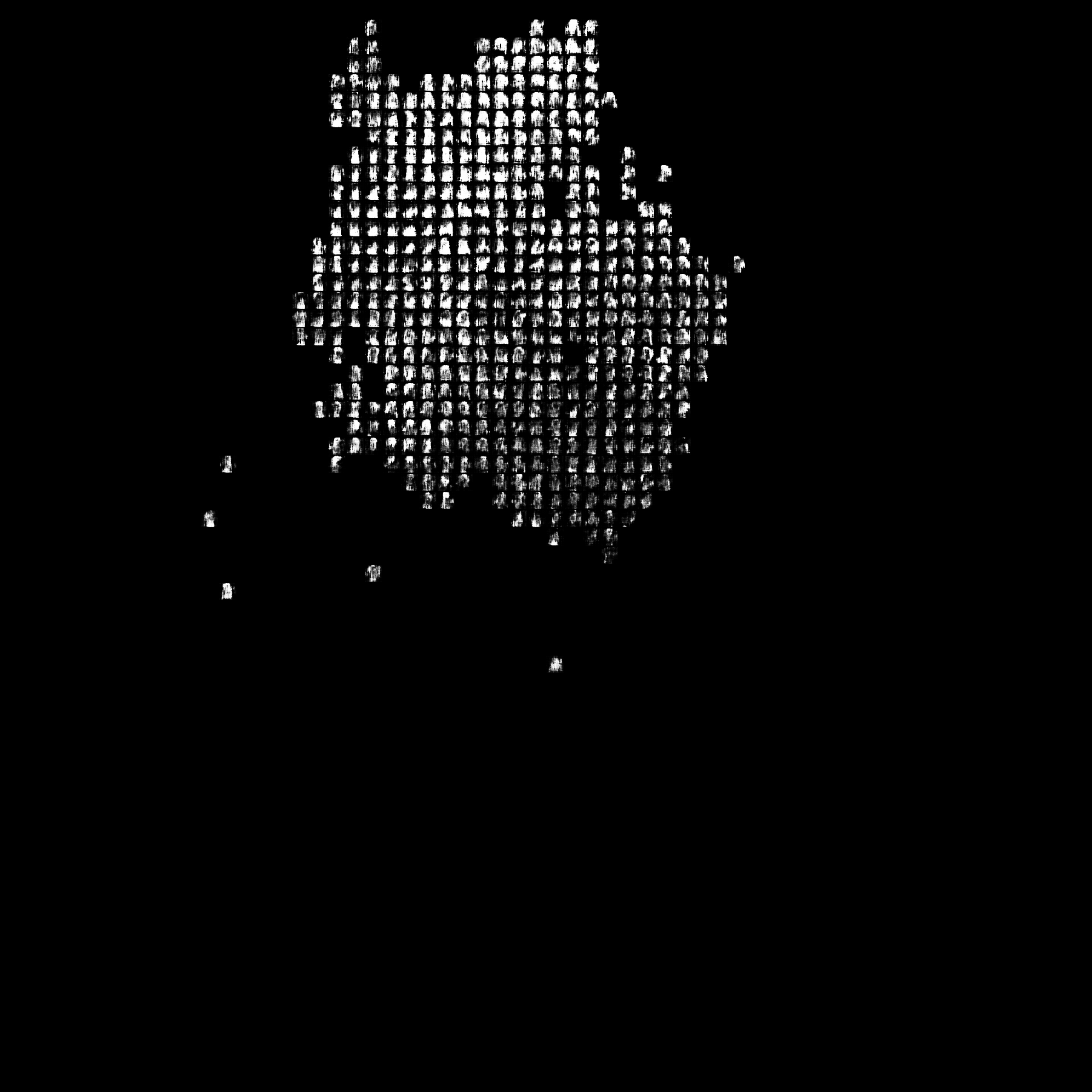}%
		\label{fig:tsne_gen_004}}
	\hfil
	\subfloat[Generation 10]{\includegraphics[width=0.37\linewidth]{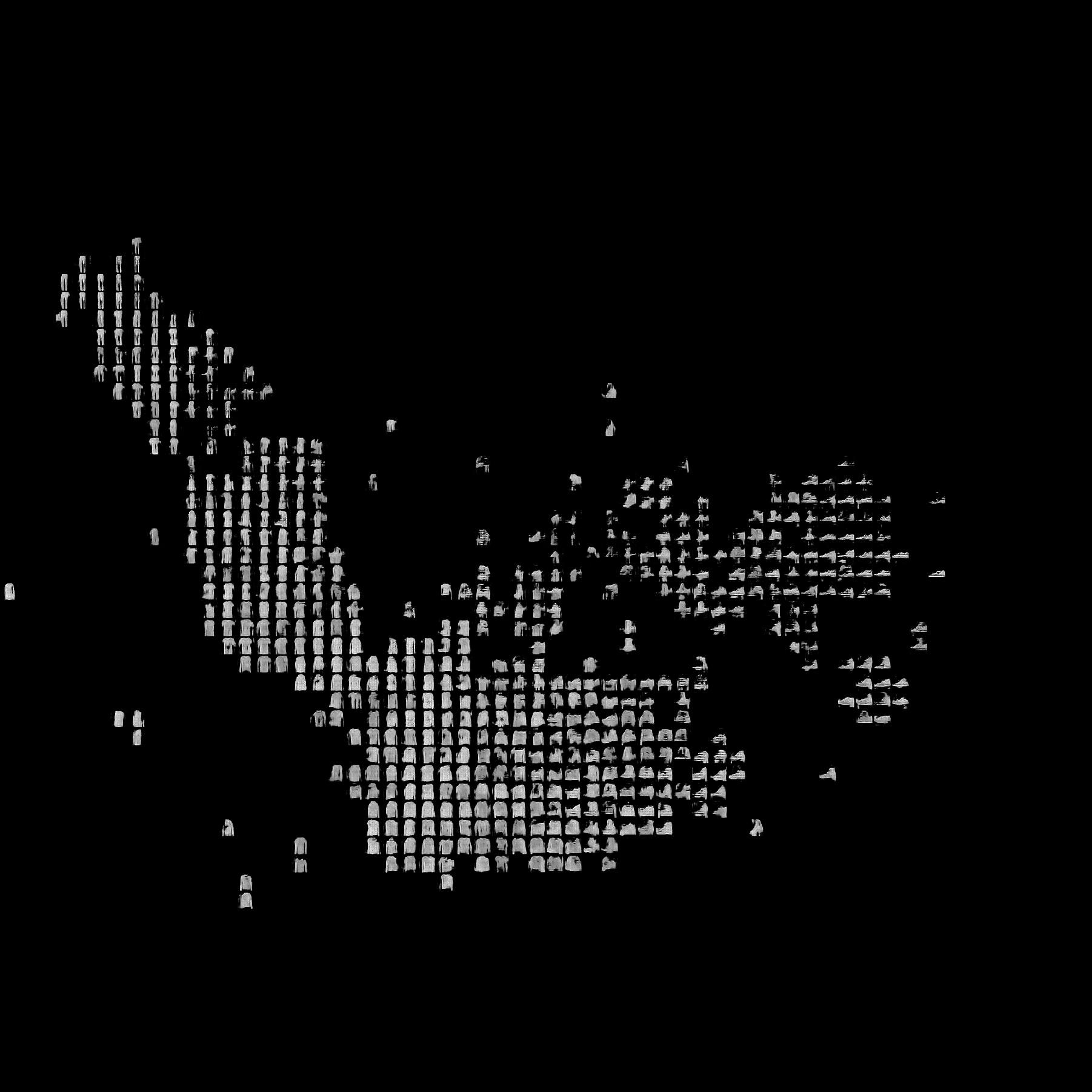}%
		\label{fig:tsne_gen_009}}
	\hfil
	\subfloat[Generation 100]{\includegraphics[width=0.37\linewidth]{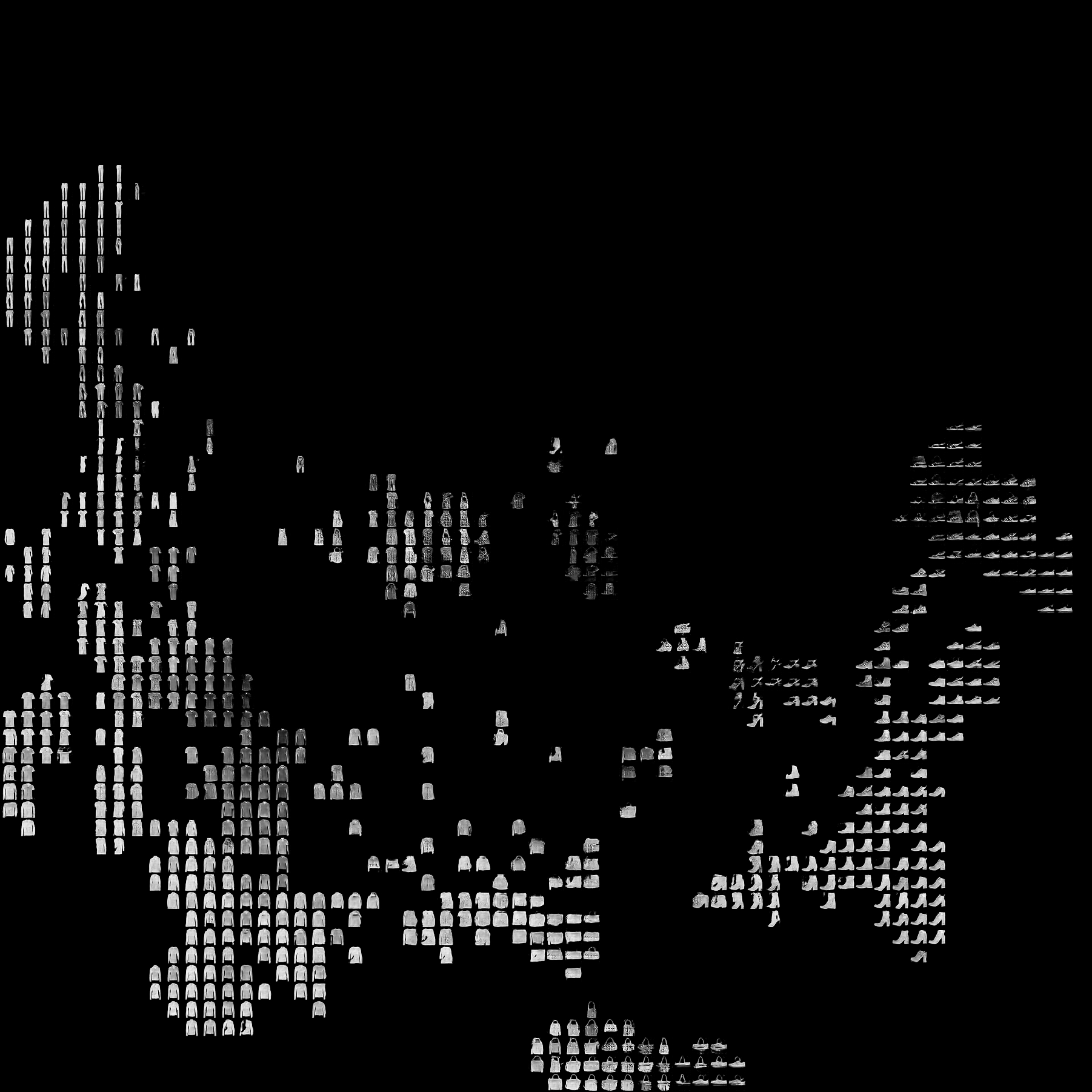}%
		\label{fig:tsne_gen_099}}
	\hfil
	\subfloat[Input Dataset]{\includegraphics[width=0.37\linewidth]{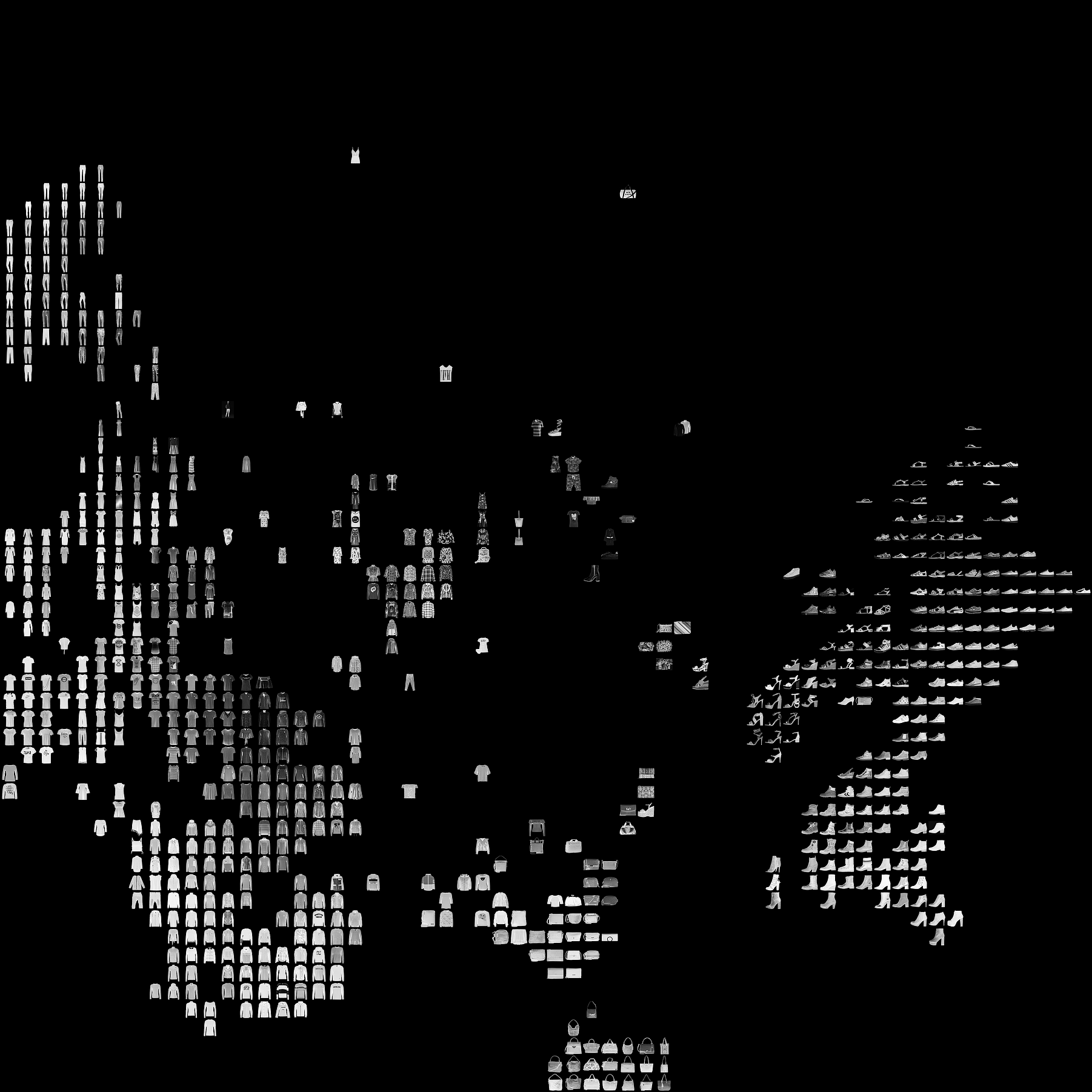}%
		\label{fig:tsne_dataset}}
	\hfil
	\caption{Two-dimensional grid revealing the distribution of images after applying t-SNE for generations 5~\subref{fig:tsne_gen_004}, 10~\subref{fig:tsne_gen_009}, 100~\subref{fig:tsne_gen_099}, and for the input dataset~\subref{fig:tsne_dataset}.}
	\label{fig:tsne}
\end{figure}

First, we show the results of a single execution of COEGAN using the parameters defined in Table~\ref{table:setup}.
Figure~\ref{fig:tsne} presents the resulting map of images after the application of t-SNE with the feature map of the best discriminator at the last generation.
In Figures~\ref{fig:tsne_gen_004}, \ref{fig:tsne_gen_009}, and \ref{fig:tsne_gen_099}, we can see the distribution of samples created by the best generators at generation $5$, $10$ and $100$, respectively.
Figure~\ref{fig:tsne_dataset} represents the distribution of the input dataset.
Initially, at generation $5$, samples are concentrated in a compact region of the grid, indicating that the distribution was not successfully captured yet.
The distribution of samples improves at generation $10$.
At the final generation, we can see that the distribution of samples is similar to the input dataset.
Therefore, Figures~\ref{fig:tsne_gen_099} and \ref{fig:tsne_dataset} presents similar structure regarding the two-dimensional grid.
Furthermore, the distribution of samples shows that the mode collapse issue does not affect the GAN model in this experiment.

We use the outcome of t-SNE to extract some metrics to quantify and represent the observations we made by visual inspection.
For this, we calculate the distances between each sample created at generations $5$, $10$, and $100$ with the samples from the input dataset.

\begin{figure}[ht]
	\centering
	\includegraphics[width=0.07\linewidth]{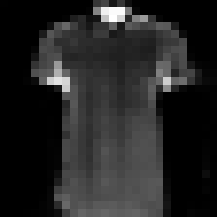}%
	\includegraphics[width=0.07\linewidth]{generator_1}%
	\includegraphics[width=0.07\linewidth]{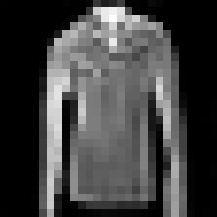}%
	\includegraphics[width=0.07\linewidth]{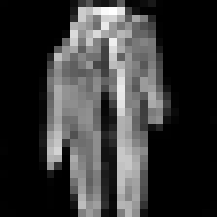}%
	\includegraphics[width=0.07\linewidth]{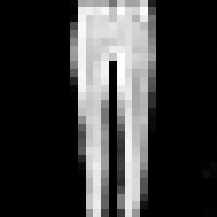}%
	\includegraphics[width=0.07\linewidth]{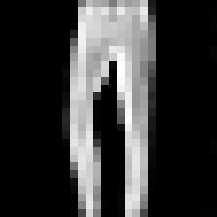}%
	\includegraphics[width=0.07\linewidth]{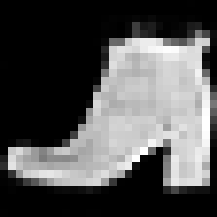}%
	\includegraphics[width=0.07\linewidth]{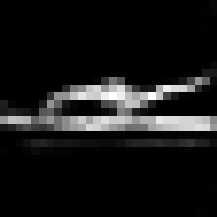}%
	\\
	\includegraphics[width=0.07\linewidth]{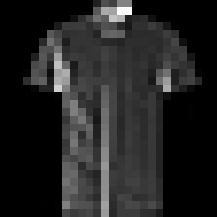}%
	\includegraphics[width=0.07\linewidth]{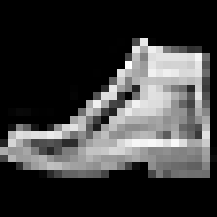}%
	\includegraphics[width=0.07\linewidth]{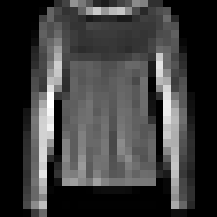}%
	\includegraphics[width=0.07\linewidth]{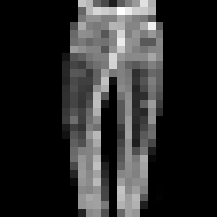}%
	\includegraphics[width=0.07\linewidth]{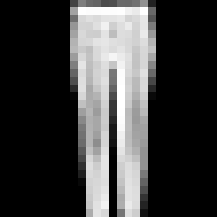}%
	\includegraphics[width=0.07\linewidth]{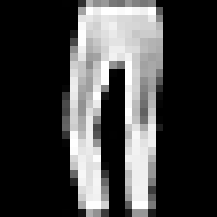}%
	\includegraphics[width=0.07\linewidth]{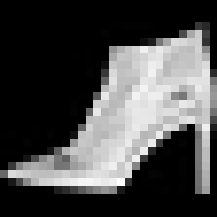}%
	\includegraphics[width=0.07\linewidth]{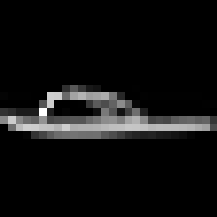}%
	\\
	\includegraphics[width=0.07\linewidth]{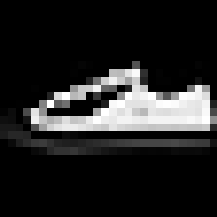}%
	\includegraphics[width=0.07\linewidth]{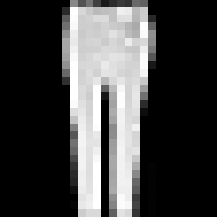}%
	\includegraphics[width=0.07\linewidth]{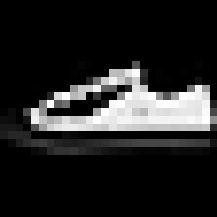}%
	\includegraphics[width=0.07\linewidth]{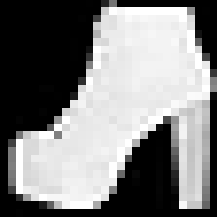}%
	\includegraphics[width=0.07\linewidth]{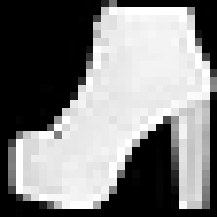}%
	\includegraphics[width=0.07\linewidth]{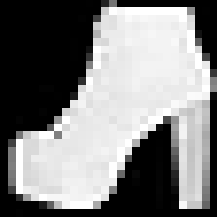}%
	\includegraphics[width=0.07\linewidth]{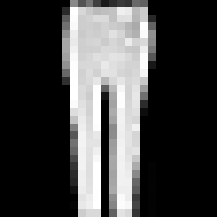}%
	\includegraphics[width=0.07\linewidth]{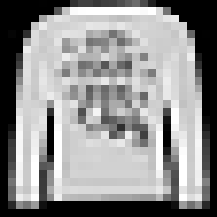}%
	\caption{Comparison of samples created by the generator and samples from the input dataset using t-SNE. First row shows samples created by generators from the last generation. Second and third rows display the nearest and farthest samples from Fashion MNIST using distances from the resulting t-SNE grid.}
	\label{fig:samples_distances}
\end{figure}

We show in Figure~\ref{fig:samples_distances} examples of created samples and their respective nearest and farthest samples from the input dataset concerning the t-SNE map.
We can see that the t-SNE map, calculated through the features trained for discriminators, is able to aggregate images based on the similarity.
Thus, the neighborhood of a sample in the t-SNE grid contains images with similar characteristics.

\begin{figure}[ht]
	\centering
	\subfloat[Histogram of distances]{\includegraphics[width=0.47\linewidth]{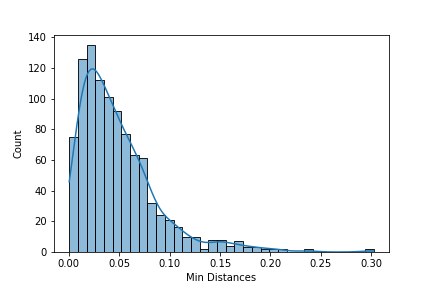}%
		\label{fig:min_distances_hist}}
	\hfil
	\subfloat[Boxplot of distances]{\includegraphics[width=0.47\linewidth]{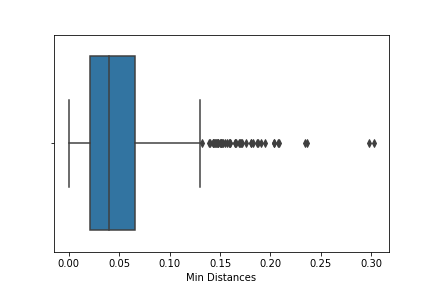}%
		\label{fig:min_distances_box}}
	\caption{Minimum distances between samples from the input dataset and samples created by generators at the last generation.}
	\label{fig:min_distances}
\end{figure}

Figure~\ref{fig:min_distances} shows the distribution of the minimum distances between each generated sample in the last generation and samples from the input dataset.
This distribution is used to calculate the threshold for the next step.
In this case, we use the value of the median ($0.0394$) as the threshold.

This threshold is applied to get the intersection between the map of the input dataset and maps of generations $5$, $10$, and $100$.
We use the number of samples in this intersection to calculate the metric to quantify the progress of the model.

\begin{figure}[h]
	\centering
	\includegraphics[width=0.47\linewidth]{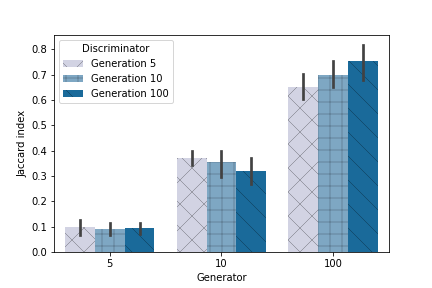}
	\caption{Average Jaccard index ($10$ executions) for generators when comparing created samples with samples drawn from the input dataset.}
	\label{fig:tsne_metric}
\end{figure}

Figure~\ref{fig:tsne_metric} shows the Jaccard index (Eq.\ref{eq:metric}) between created samples and the input dataset for generators and discriminators at generations $5$, $10$, and $100$ for ten executions.
The results evidenced that discriminators in all three generations were able to identify poor samples produced by generators in generation $5$.
This is evidenced by the low Jaccard index, revealing that samples do not have strong similarities with the input dataset.
For samples created at generation $10$, we can see that more evolved discriminators are slightly more capable of identifying fake samples.
Finally, all three discriminators were able to successfully distribute samples when evaluated with samples created at the final generation.
Besides, more evolved discriminators are better on the distribution of samples, leading to better results concerning the proposed metric.
As expected, the metric for generators in the last generation is smaller than $1$ ($0.753\pm0.112$), evidencing that generators are not only capable of capturing the input distribution but also to produce innovative samples.

\section{Conclusions}~\label{sec:conclusions}
In this paper, we propose a new evaluation method to assess the progress of generators and discriminators in Generative Adversarial Networks (GANs).
For this, we propose a method based on t-SNE to visually inspect the quality of discriminators and generators in GANs.
Furthermore, a metric based on the Jaccard index between t-SNE maps was designed to quantitatively represent the aspects of the model.

This evaluation method was applied to a GAN trained by an Evolutionary Algorithm to validate our proposal.
COEGAN combines competitive coevolution and neuroevolution on the evolution of GANs and is capable of avoiding stability issues on training, using selective pressure to guide the progress of generators and discriminators.
Therefore, we use this model in experiments to show the evolution of discriminators and generators through our evaluation method, providing further evidence of the evolutionary aspects of COEGAN.

Results demonstrate both by visual inspection and the proposed metric that COEGAN is able to gradually evolve GANs, avoiding problems such as mode collapse.
We also show that the use of t-SNE proposed in this work can aggregate similar samples and provide their efficient distribution in a two-dimensional grid.

For future work, we intend to expand the experiments to analyze the results in more complex datasets, such as CelebA and CIFAR10.
Furthermore, we expect to incorporate new mechanisms proposed for GANs into an Evolutionary Algorithm to assess the contributions of them when evaluated with our method.

\section*{Acknowledgments}\label{sec:acknowledgments}
This work is partially funded by the project grant DSAIPA/DS/0022/2018 (GADgET), by national funds through the FCT - Foundation for Science and Technology, I.P., within the scope of the project CISUC - UID/CEC/00326/2020 and by European Social Fund, through the Regional Operational Program Centro 2020. We also thank the NVIDIA Corporation for the hardware granted to this research.

\bibliographystyle{splncs04}
\bibliography{costa}

\begin{thebibliography}{10}
\providecommand{\url}[1]{\texttt{#1}}
\providecommand{\urlprefix}{URL }
\providecommand{\doi}[1]{https://doi.org/#1}

\bibitem{al2018towards}
Al-Dujaili, A., Schmiedlechner, T., Hemberg, E., O’Reilly, U.M.: Towards
  distributed coevolutionary {GANs}. In: AAAI 2018 Fall Symposium (2018)

\bibitem{arjovsky2017wasserstein}
Arjovsky, M., Chintala, S., Bottou, L.: Wasserstein generative adversarial
  networks. In: International Conference on Machine Learning. pp. 214--223
  (2017)

\bibitem{assunccao2019denser}
Assun{\c{c}}{\~a}o, F., Louren{\c{c}}o, N., Machado, P., Ribeiro, B.: {DENSER}:
  Deep evolutionary network structured representation. Genetic Programming and
  Evolvable Machines  \textbf{20}(1),  5--35 (2019)

\bibitem{belkina2019automated}
Belkina, A.C., Ciccolella, C.O., Anno, R., Halpert, R., Spidlen, J.,
  Snyder-Cappione, J.E.: Automated optimized parameters for t-distributed
  stochastic neighbor embedding improve visualization and analysis of large
  datasets. Nature communications  \textbf{10}(1),  1--12 (2019)

\bibitem{berthelot2017began}
Berthelot, D., Schumm, T., Metz, L.: {BEGAN}: Boundary equilibrium generative
  adversarial networks. arXiv preprint arXiv:1703.10717  (2017)

\bibitem{brock2018large}
Brock, A., Donahue, J., Simonyan, K.: Large scale {GAN} training for high
  fidelity natural image synthesis. In: International Conference on Learning
  Representations (2019)

\bibitem{costa2019evaluating}
Costa, V., Louren{\c{c}}o, N., Correia, J., Machado, P.: {COEGAN}: Evaluating
  the coevolution effect in generative adversarial networks. In: Proceedings of
  the Genetic and Evolutionary Computation Conference. pp. 374--382. ACM (2019)

\bibitem{costa2020neuroevolution}
Costa, V., Louren{\c{c}}o, N., Correia, J., Machado, P.: Neuroevolution of
  generative adversarial networks. In: Deep Neural Evolution, pp. 293--322.
  Springer (2020)

\bibitem{costa2019coevolution}
Costa, V., Louren{\c{c}}o, N., Machado, P.: Coevolution of generative
  adversarial networks. In: International Conference on the Applications of
  Evolutionary Computation (Part of EvoStar). pp. 473--487. Springer (2019)

\bibitem{fedus2017many}
Fedus, W., Rosca, M., Lakshminarayanan, B., Dai, A.M., Mohamed, S., Goodfellow,
  I.: Many paths to equilibrium: {GAN}s do not need to decrease a divergence at
  every step. In: International Conference on Learning Representations (2018)

\bibitem{garciarena2018evolved}
Garciarena, U., Santana, R., Mendiburu, A.: Evolved {GANs} for generating
  pareto set approximations. In: Proceedings of the Genetic and Evolutionary
  Computation Conference. pp. 434--441. GECCO '18, ACM, New York, NY, USA
  (2018)

\bibitem{NIPS2014_5423}
Goodfellow, I., Pouget-Abadie, J., Mirza, M., Xu, B., Warde-Farley, D., Ozair,
  S., Courville, A., Bengio, Y.: Generative adversarial nets. In: NIPS. Curran
  Associates, Inc. (2014)

\bibitem{gulrajani2017improved}
Gulrajani, I., Ahmed, F., Arjovsky, M., Dumoulin, V., Courville, A.C.: Improved
  training of wasserstein {GANs}. In: Advances in Neural Information Processing
  Systems. pp. 5769--5779 (2017)

\bibitem{heusel2017gans}
Heusel, M., Ramsauer, H., Unterthiner, T., Nessler, B., Hochreiter, S.: {GANs}
  trained by a two time-scale update rule converge to a local nash equilibrium.
  In: Advances in Neural Information Processing Systems. pp. 6629--6640 (2017)

\bibitem{karras2018progressive}
Karras, T., Aila, T., Laine, S., Lehtinen, J.: Progressive growing of {GAN}s
  for improved quality, stability, and variation. In: International Conference
  on Learning Representations (2018)

\bibitem{karras2018style}
Karras, T., Laine, S., Aila, T.: A style-based generator architecture for
  generative adversarial networks. arXiv preprint arXiv:1812.04948  (2018)

\bibitem{kingma2015adam}
Kingma, D.P., Ba, J.: Adam: A method for stochastic optimization. In:
  International Conference on Learning Representations (ICLR) (2015)

\bibitem{kobak2019art}
Kobak, D., Berens, P.: The art of using t-{SNE} for single-cell
  transcriptomics. Nature communications  \textbf{10}(1),  1--14 (2019)

\bibitem{lourencco2015sge}
Louren{\c{c}}o, N., Pereira, F.B., Costa, E.: {SGE}: A structured
  representation for grammatical evolution. In: International Conference on
  Artificial Evolution (Evolution Artificielle). pp. 136--148. Springer (2015)

\bibitem{maaten2008visualizing}
Maaten, L.v.d., Hinton, G.: Visualizing data using t-{SNE}. Journal of machine
  learning research  \textbf{9}(Nov),  2579--2605 (2008)

\bibitem{mao2017least}
Mao, X., Li, Q., Xie, H., Lau, R.Y., Wang, Z., Smolley, S.P.: Least squares
  generative adversarial networks. In: 2017 IEEE International Conference on
  Computer Vision (ICCV). pp. 2813--2821. IEEE (2017)

\bibitem{miikkulainen2017evolving}
Miikkulainen, R., Liang, J., Meyerson, E., Rawal, A., Fink, D., Francon, O.,
  Raju, B., Navruzyan, A., Duffy, N., Hodjat, B.: Evolving deep neural
  networks. arXiv preprint arXiv:1703.00548  (2017)

\bibitem{miyato2018spectral}
Miyato, T., Kataoka, T., Koyama, M., Yoshida, Y.: Spectral normalization for
  generative adversarial networks. In: International Conference on Learning
  Representations (2018)

\bibitem{radford2015unsupervised}
Radford, A., Metz, L., Chintala, S.: Unsupervised representation learning with
  deep convolutional generative adversarial networks. arXiv preprint
  arXiv:1511.06434  (2015)

\bibitem{russakovsky2015imagenet}
Russakovsky, O., Deng, J., Su, H., Krause, J., Satheesh, S., Ma, S., Huang, Z.,
  Karpathy, A., Khosla, A., Bernstein, M., et~al.: Imagenet large scale visual
  recognition challenge. International Journal of Computer Vision
  \textbf{115}(3),  211--252 (2015)

\bibitem{salimans2016improved}
Salimans, T., Goodfellow, I., Zaremba, W., Cheung, V., Radford, A., Chen, X.:
  Improved techniques for training {GANs}. In: Advances in Neural Information
  Processing Systems. pp. 2234--2242 (2016)

\bibitem{sims1994evolving}
Sims, K.: Evolving 3d morphology and behavior by competition. Artificial life
  \textbf{1}(4),  353--372 (1994)

\bibitem{neat}
Stanley, K.O., Miikkulainen, R.: Evolving neural networks through augmenting
  topologies. Evolutionary computation  \textbf{10}(2),  99--127 (2002)

\bibitem{szegedy2016rethinking}
Szegedy, C., Vanhoucke, V., Ioffe, S., Shlens, J., Wojna, Z.: Rethinking the
  inception architecture for computer vision. In: Proceedings of the IEEE
  Conference on Computer Vision and Pattern Recognition. pp. 2818--2826 (2016)

\bibitem{toutouh2019spatial}
Toutouh, J., Hemberg, E., O’Reilly, U.M.: Spatial evolutionary generative
  adversarial networks. arXiv preprint arXiv:1905.12702  (2019)

\bibitem{wang2018evolutionary}
Wang, C., Xu, C., Yao, X., Tao, D.: Evolutionary generative adversarial
  networks. arXiv preprint arXiv:1803.00657  (2018)

\bibitem{xiao2017fashion}
Xiao, H., Rasul, K., Vollgraf, R.: Fashion-{MNIST}: A novel image dataset for
  benchmarking machine learning algorithms. arXiv preprint arXiv:1708.07747
  (2017)

\bibitem{yao1999evolving}
Yao, X.: Evolving artificial neural networks. Proceedings of the IEEE
  \textbf{87}(9),  1423--1447 (1999)

\bibitem{zhang2018self}
Zhang, H., Goodfellow, I., Metaxas, D., Odena, A.: Self-attention generative
  adversarial networks. arXiv preprint arXiv:1805.08318  (2018)

\bibitem{zhang2018stackgan++}
Zhang, H., Xu, T., Li, H., Zhang, S., Wang, X., Huang, X., Metaxas, D.N.:
  {StackGAN++}: Realistic image synthesis with stacked generative adversarial
  networks. IEEE Transactions on Pattern Analysis and Machine Intelligence
  \textbf{41}(8),  1947--1962 (2018)

\end{thebibliography}

\end{document}